\documentclass{ecai} 

\usepackage{latexsym}
\usepackage{amssymb}
\usepackage{amsmath}
\usepackage{amsthm}
\usepackage{booktabs}
\usepackage{colortbl} 
\usepackage{enumitem}
\usepackage{graphicx}
\usepackage{color}
\usepackage{multirow}
\usepackage{caption}
\usepackage[table,xcdraw]{xcolor}

\newcommand{\BibTeX}{B\kern-.05em{\sc i\kern-.025em b}\kern-.08em\TeX}

\begin{document}


\begin{frontmatter}


\paperid{7055} 


\title{MemoCue: Empowering LLM-Based Agents for \\ Human Memory Recall via Strategy-Guided Querying}


\author[A]{\fnms{Qian}~\snm{Zhao}}
\author[A]{\fnms{Zhuo}~\snm{Sun}}
\author[A]{\fnms{Bin}~\snm{Guo}}
\author[A]{\fnms{Zhiwen}~\snm{Yu}}

\address[A]{Northwestern Polytechnical University}


\begin{abstract}
Agent-assisted memory recall is one critical research problem in the field of human-computer interaction. In conventional methods, the agent can retrieve information from its equipped memory module to help the person recall incomplete or vague memories. The limited size of memory module hinders the acquisition of complete memories and impacts the memory recall performance in practice. Memory theories suggest that the person's relevant memory can be proactively activated through some effective cues. Inspired by this, we propose a novel strategy-guided agent-assisted memory recall method, allowing the agent to transform an original query into a cue-rich one via the judiciously designed strategy to help the person recall memories. To this end, there are two key challenges. (1) How to choose the appropriate recall strategy for diverse forgetting scenarios with distinct memory-recall characteristics? (2) How to obtain the high-quality responses leveraging recall strategies, given only abstract and sparsely annotated strategy patterns? To address the challenges, we propose a Recall Router framework. Specifically, we design a 5W Recall Map to classify memory queries into five typical scenarios and define fifteen recall strategy patterns across the corresponding scenarios. We then propose a hierarchical recall tree combined with the Monte Carlo Tree Search algorithm to optimize the selection of strategy and the generation of strategy responses. We construct an instruction tuning dataset and fine-tune multiple open-source large language models (LLMs) to develop MemoCue, an agent that excels in providing memory-inspired responses. Experiments on three representative datasets show that MemoCue surpasses LLM-based methods by 17.74\% in recall inspiration. Further human evaluation highlights its advantages in memory-recall applications. 

\end{abstract}

\end{frontmatter}


\section{Introduction}

Tip-of-the-Tongue (TOT)~\cite{brown1966tip} is a phenomenon that usually occurs in our daily life, where one person has the memory but struggles to recall it immediately. As the pace of society accelerates, this issue becomes increasingly serious, and affects the quality of life. For example, when you need keys but can't remember where you left them, it can be frustrated and helpless. Therefore, finding a solution to alleviate this issue becomes an urgent need. 
\begin{figure}[h]
    \centering
    \includegraphics[width=\linewidth]{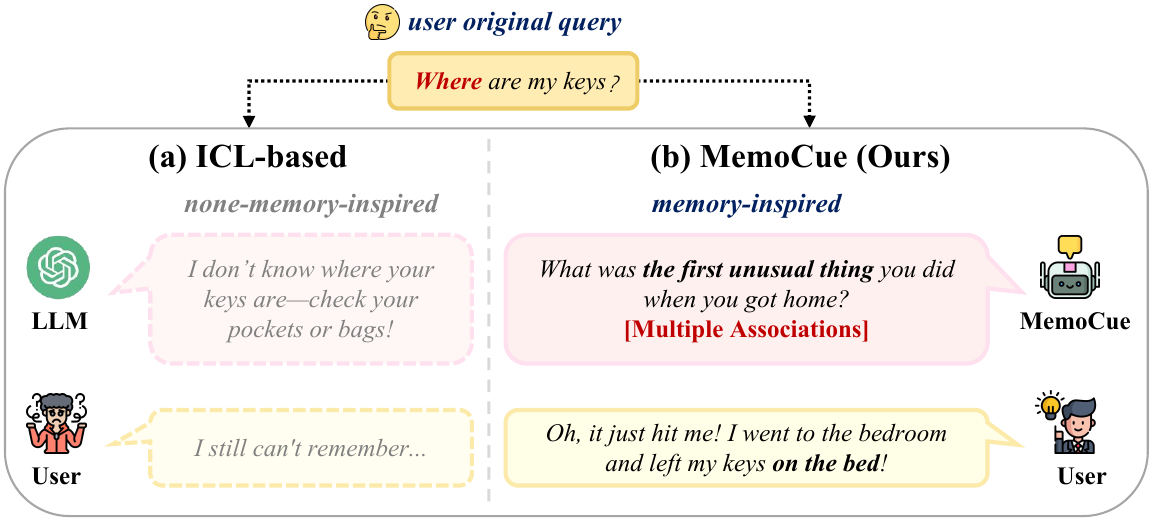}
    \captionsetup{justification=justified, singlelinecheck=false, skip=6pt,belowskip=5pt}
    \caption{A comparison of forgetting scenario execution with traditional LLM method and our MemoCue agent. 
    (a) Ineffective responses generated by LLM using zero-shot CoT prompting, one In-Context Learning (ICL) method. (b) Strategy-guided responses generated by MemoCue agent, which transforms an original query into a cue-rich one, thereby offering useful memory-inspired information to the user. The recall strategy used is marked in red color. }
    \label{fig:5w}
\end{figure}

With the advancements in the study of human cognition and memory~\cite{chandra2025episodic}, some studies have attempted to assist human memory recall through human-computer interaction. HippoCamera~\cite{martin2022smartphone} is a smartphone-based memory recall system that helps reactivate memories by replaying memory cues from daily life. \citeauthor{schindler2022encoding} (\citeyear{schindler2022encoding}) developed a software to enhance the memory retention by using the social feedback, called interactional behavioral system. \citeauthor{georgiev2021virtual} (\citeyear{georgiev2021virtual}) explored different types of virtual reality applications to improve the working memory~\cite{baddeley1992working} through immersive experience. These methods mainly rely on the explicit recording and the repeated playback of past experiences to assist the person to recall. The relationship among memory fragments is fixed and it lacks the evolution with adding new memory fragments. The rise of agents equipped with memory modules~\cite{zhang2024survey} and large language models (LLMs)~\cite{openai2023gpt} presents a promising solution to alleviate the issue. It can perceive the user's memory context and respond to the user's query by employing retrieval-augmented generation (RAG) techniques~\cite{sarthiraptor}. HippoRAG~\cite{gutierrez2024hipporag} integrated external information continuously inspired by the neocortex and hippocampus in human memory. LongMem~\cite{wang2023augmenting} introduced a decoupled network architecture designed for storing and updating long contexts for agent retrieval. While these methods achieve the dynamic organization of stored memory fragments, they only rely on the passive retrieval of the agent's stored memories to assist the person to recall. The recall performance is determined by the amount of stored memories. 

However, when agents are applied to the real world, it is very challenging to acquire and store the entire memory data. First, the acquisition of memory data is limited by temporal and spatial constraints. Real-world memories arise from large-scale and multimodal data sources, such as text~\cite{zhong2024memorybank}, video~\cite{song2024moviechat}, which are often unpredictable in time and space. Furthermore, data collection relies heavily on the trust from users~\cite{kandappu2021privacyprimer} in terms of legal and privacy considerations. Second, the limited storage space of the agent makes it difficult to store such a huge amount of memory information from the users over a large span of time and space. Thus, the incomplete memory data significantly restricts the agent's ability to help human recall.

\begin{figure*}[htbp]
  \centering
  \includegraphics[width=0.88\linewidth]{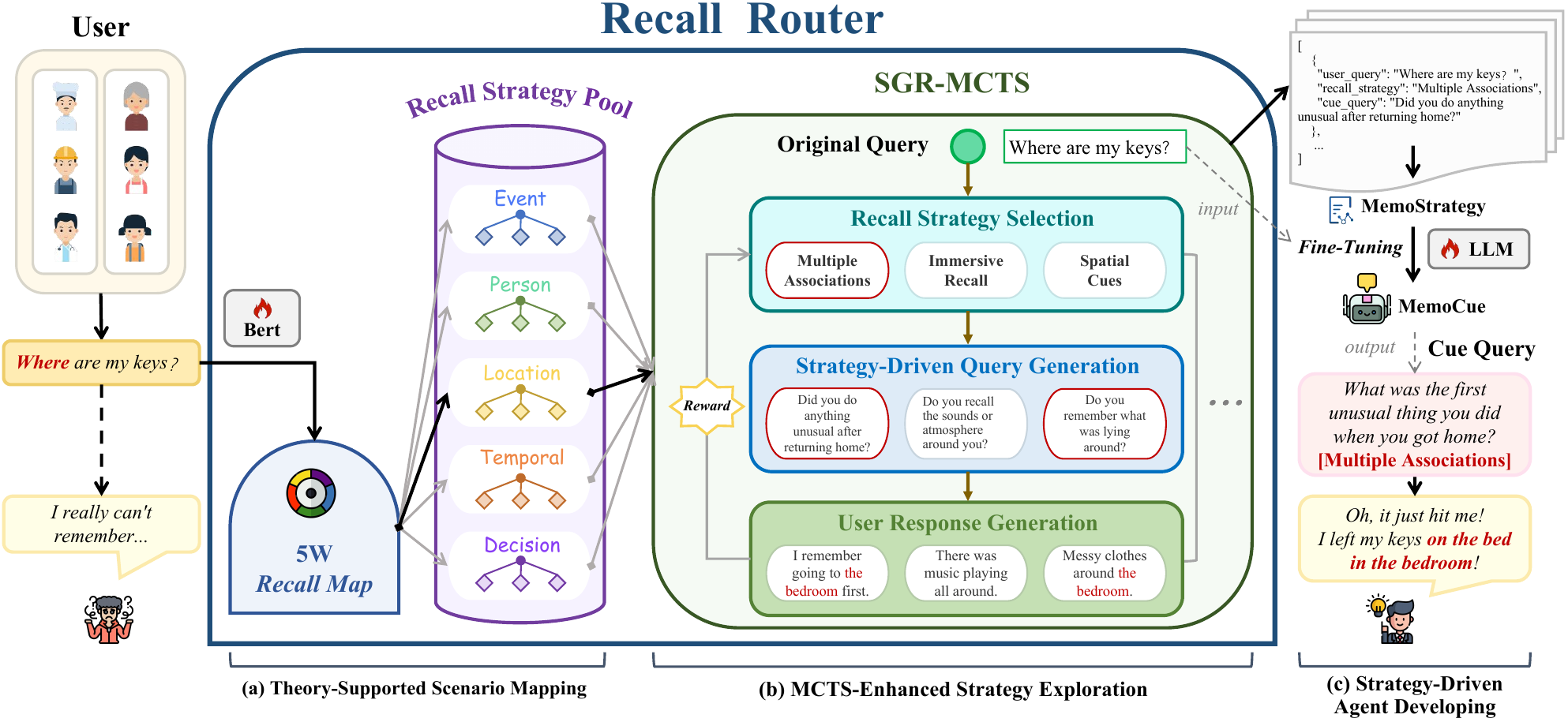}
  \captionsetup{justification=justified, singlelinecheck=false, skip=6pt,belowskip=5pt}
  \caption{A schematic diagram of the overall memory-recall architecture for SGR problem. }
  \label{fig:overview}
\end{figure*}

Inspired by the discovery that human forgetfulness is often caused by the ineffective activation of memories rather than the memory loss~\cite{collins1975spreading,meyer1992tip}, we formulate a novel \textbf{S}trategy-\textbf{G}uided \textbf{R}ecall \textbf{(SGR)} problem to achieve the agent-assisted proactive memory recall. Different from the traditional solutions, the agent can strategically guide the user to find associations across diverse memories and activate some memories progressively and achieve the memory recall in SGR, thereby achieving the memory recall. With the guidance strategy, this can alleviate the heavy reliance of memory recall performance on the amount of acquired and stored memories. However, there are various types of memory queries from the user in practice and it requires the unique recall strategy for each memory query type based on its characteristics. Moreover, it is necessary for the recall strategy to be adaptively adjusted to the user's responses. In addition, the dataset for the SGR problem should include the diverse recall strategies and the appropriate evaluation metrics, which is missing. While the recent LLMs have shown remarkable capabilities in response generation across a wide range of open-domain tasks, their ability to effectively guide human memory recall is still limited. As shown in Figure~\ref{fig:5w}, the response generated by LLM using ICL method exhibits a lack of specificity and recall cues.

To address the aforementioned challenges, we propose a Recall Router framework, as shown in Figure~\ref{fig:overview}. This framework consists of a 5W based scenario query classification and a Monte Carlo Tree Search (MCTS)~\cite{kocsis2006bandit} based strategy response generation. The 5W based scenario classification uses the 5W \textit{(What, Who, Where, When, and Why)} recall map to classify the user queries into five scenarios: Event, Person, Location, Temporal, and Decision. Based on memory theory~\cite{herz1997effects,eichenbaum1999hippocampus}, we define a Recall Strategy Pool for the corresponding scenarios, which includes fifteen potential recall strategy patterns, as illustrated in Figure~\ref{fig:tree}. We propose SGR-MCTS to generate high-quality recall strategy responses, where a fine-grained reward mechanism is designed to evaluate the effect of memory-recall guidance. To validate the effectiveness of proposed Recall Router, we construct MemoStrategy, an instruction tuning recall strategy dataset. At last, by fine-tuning on open-source LLM, we develop MemoCue, an LLM-based agent capable of generating strategy-driven cue queries and providing high-quality recall guidance. 

Our contributions can be summarized as follows: 
\begin{itemize}
    \item We propose a Recall Router framework for the formulated SGR problem, aiming to exploit the capacity of LLM-based agents to generate memory-inspired responses and strategically guide the user to activate some memories progressively. 
    \item We design a 5W Recall Map, which classifies user queries to five typical forgetting scenarios, and defines the corresponding recall strategy patterns. Then we propose a SGR-MCTS algorithm to generate effective recall strategy responses, where a fine-grained reward mechanism is designed to evaluate the effect of memory-recall guidance. 
    \item We construct an instruction tuning dataset and develop an LLM-based recall guidance agent MemoCue, which is able to generate strategy-driven cue queries and provide high-quality recall guidance. To evaluate reasonably, we design an evaluation metric which balances the query novelty and response accuracy. 
    \item We conduct extensive experiments on three public memory datasets to validate the effectiveness of MemoCue. To evaluate reasonably, we design an evaluation metric which balances the query novelty and response accuracy. The results demonstrate the superiority of MemoCue in memory recall compared to baselines. 
\end{itemize}

\section{Related Work}
Human memory, as a fundamental component of human cognition, has long been a topic across various fields~\cite{sarfraz2023sparse,gutierrez2024hipporag}. The process of human memory is divided into three stages: encoding, storage, and retrieval, with the retrieval stage refers to the memory recall~\cite{squire1986mechanisms}. Tip-of-the-Tongue (TOT) is frustrated phenomenon which refers to the situation that an individual unable to recall a certain memory immediately even though he know it. Actually, the phenomenon is not only a common occurrence in daily life but also a long-term research focus in cognitive psychology and neuroscience ~\cite{burke1991tip}.

Previous studies have made some attempts to address the problem and improve human memory recall from the perspective of human-computer interaction~\cite{martin2022smartphone,schindler2022encoding,georgiev2021virtual}. However, these approaches primarily rely on the active recording and repetition of key memories, which fails to account for the unpredictability of forgetting scenarios. Furthermore, their reliance on stored data limits their capacity to guide users actively through the memory recall process.

Recently, LLM-based agent applications gain widespread attention in various fields, such as personal assistant~\cite{lu2023memochat}, social simulation~\cite{li2023metaagents,kaiya2023lyfe}, etc. However, the application of LLM-based agent in assisting human memory recall remains insufficiently addressed due to limited memory data. For example, for smart home agents, only partial event summaries can be stored due to
privacy or limited device memory~\cite{khan2022human}. 
Many prominent memory theories~\cite{craik1972levels,eichenbaum1999hippocampus} suggest that forgetfulness arises from a lack effective cues to activate relevant memories, indicating that guiding users to activate key memories is more universally effective than passively relying on incomplete historical memory. 

Large language models (LLMs) empowered by in-context learning (ICL) demonstrate remarkable capabilities in generating coherent and contextually relevant responses, even without task-specific fine-tuning~\cite{dong2022survey,naveed2023comprehensive}. However, they often produce generic or irrelevant responses, as they lack tailored guidance or context-sensitive strategies that are essential for effective memory activation. This limitation becomes particularly evident when users struggle with vague or fragmented memories. Our work, however, considers the cognitive characteristics of human in the memory recall process, enabling agents to offer more targeted and cognitively aligned support in human memory recall, thereby providing a useful assistance and a promising direction for the application of agents in human-centered tasks.


\section{Methodology}
\subsection{Overview}
To alleviate the serious dependence of memory recall performance on the amount of stored memories, we formulate the \textbf{S}trategy-\textbf{G}uided \textbf{R}ecall \textbf{(SGR)} problem, which aims to strategically transform the user's original query into the memory-inspired cue query and guide the user to activate some memory progressively. For example, given an user original query \(Q_u\), such as \textit{``Where are my keys?''}, it can be transformed into a cue query \(Q_c\) with strategy \(s_i\), such as \textit{``Did you do anything unusual after returning home?''}, utilizing the \textit{``Multiple Associations''} strategy. The mapping process can be expressed as Equation~\ref{eq:1}, where \(S = \{s_1, s_2, \dots, s_n\}\) is the strategy set.   
\begin{equation}
  \label{eq:1}
  Q_c = \mathcal{T} \left ( Q_u, S \right )
\end{equation}

The essence of solving SGR problem is to optimize the recall strategy selection for the user's query. To achieve this, we propose a Recall Router framework, which mainly consists of the 5W Recall Map and the SGR-MCTS algorithm, as shown in Figure~\ref{fig:overview}. For the user's memory query, the whole memory recall process includes three major steps, where the first two steps are executed in the Recall Router framework: (1) The Theory-Supported Scenario Mapping \textbf{(Section 4.2)} includes the 5W Recall Map and the Recall Strategy Pool. Specifically, when the user query \textit{(none-memory-inspired)} arrives, it is first classified based on its semantic meaning using the fine-tuned RoBERTa model~\cite{liu2019roberta}. Details are shown in Appendix B. According to the classified forgetting scenario, the corresponding strategy pattern will be selected from the Recall Strategy Pool. (2) The MCTS-Enhanced Strategy Exploration \textbf{(Section 4.3)} is based on SGR-MCTS algorithm, which includes Recall Tree Construction and Recall Tree Traversal. The former is devided into the high-level strategy selection and the low-level cue query generation, and the latter consists of four key parts: Selection, Expansion, Simulation, and Backpropagation. (3) The Strategy-Driven Agent Developing \textbf{(Section 4.4)} is implemented by fine-tuning LLM with MemoStrategy dataset, which is constructed from the previous two steps (i.e., Recall Router) . The developed agent is called MemoCue and capable of generating cue queries \textit{(memory-inspired)} to help the user recall memory.  

\subsection{Theory-Supported Scenario Mapping}
Inspired by Lasswell's ``5W'' model of communication~\cite{lasswell1948structure}, we propose the 5W Recall Map for memory recall, where the ``5W'' refers to What, Who, Where, When, and Why. These correspond to five typical forgetting scenarios: Event, Person, Location, Temporal, and Decision, as shown in Figure~\ref{fig:5w}. Our model classifies user queries based on their type of forgetfulness, identifies recall difficulties, and guides them using appropriate memory-recall strategies. Details of definitions of strategies can be found in Appendix A.

\subsection{MCTS-Enhanced Strategy Exploration}
For the strategy exploration in SGR problem, we divide it into the high-level strategy selection and the low-level cue query generation. By modeling it as a Hierarchical Markov Decision Process (HMDP), we propose SGR-MCTS to solve it. The process of SGR-MCTS includes recall tree construction and recall tree traversal, which will be presented in the following. Different from the conventional MCTS algorithm, we construct a hierarchical recall tree and design a fine-grained reward mechanism for SGR-MCTS based on the simulated user feedback. The goal is to obtain the high-quality strategy responses that can guide users to activate memories better.

\subsubsection{Recall Tree Construction}
The recall tree consists of nodes and edges, where nodes represent the current dialogue state and edges represent the current strategy actions. Based on the characteristics of SGR problem, the recall tree is divided into the high-level recall strategy selection and the low-level cue query generation. 

\begin{figure*}[htbp]
  \includegraphics[width=0.9\linewidth]{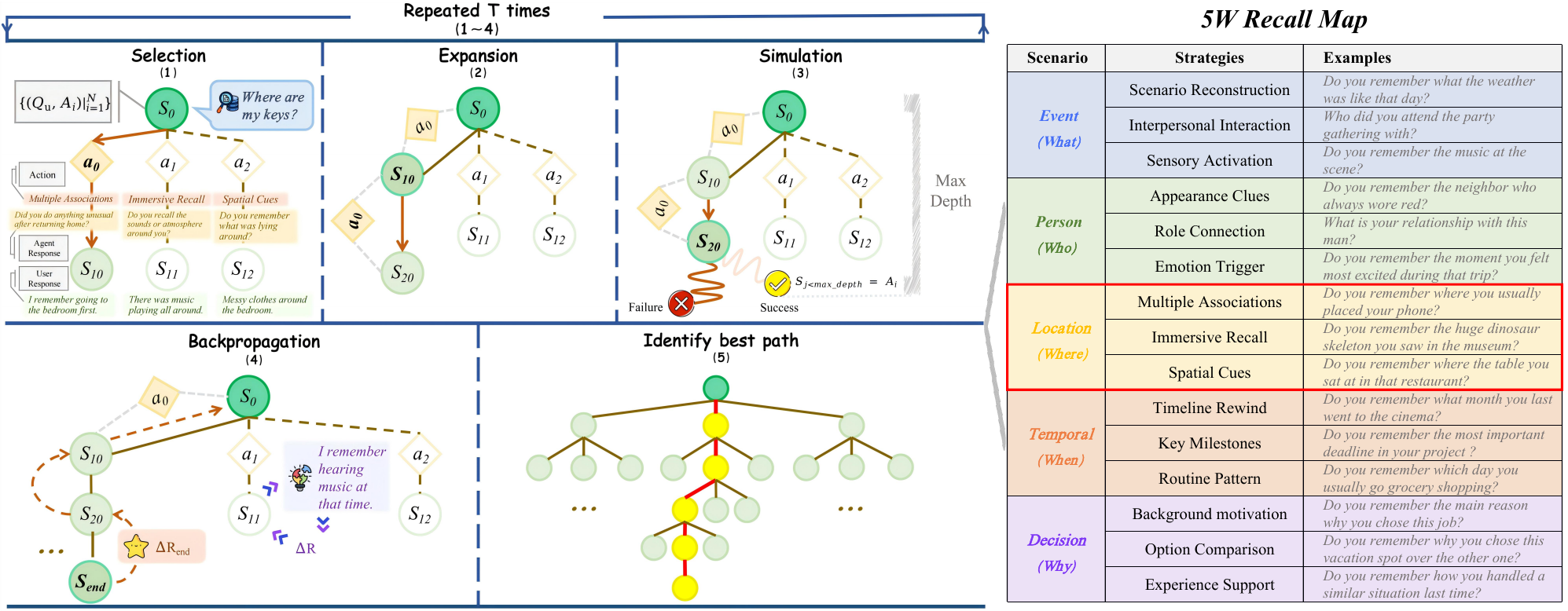}
  \centering
  \captionsetup{justification=justified, singlelinecheck=false, skip=6pt,belowskip=5pt}
  \caption{Overview of our proposed SGR-MCTS for memory-recall strategy responses generating. There are four key steps: Selection, Expansion, Simulation, and Backpropagation. Examples illustrating the 5W recall map including five typical forgetting scenarios and corresponding fifteen recall strategies are shown in the right, which constitutes the action space. We perform \(T\) iterations and collect the strategy response set on the top-\(k\) path for each query. }
  \label{fig:tree}
\end{figure*}

\paragraph{High Level: Recall Strategy Selection.}we define the high-level state space \(S^{h}\) as a current context of SGR. The high-level state \(s^{h}\) can be expressed as Equation~\ref{eq:2}. 
\begin{equation}
  \label{eq:2}
  s_{t}^{h}  = \left ( Q_u, H_t, M_t \right )
\end{equation}
where \(Q_u\) is the user original query, \(H_t\) is dialog history, including the strategies used, the generated cue queries, and user responses, and \(M_t\) is the relevant memory retrieved from user memory bank up to turn \(t\). 

The high-level action space \(\mathcal{A}^{h}\) consists of 15 recall strategy patterns, which we construct under five typical forgetting scenarios based on human memory theory. The strategy categories and examples are shown in Figure~\ref{fig:tree}. An action \(a_{t}^{h}\) \((a_{t}^{h}  \in \mathcal{A}^{h}) \) represents a specific recall strategy pattern, which aims to maximize the reward \(R^{h} ( s_{t}^{h}, a_{t}^{h}  )\) based on the feedback from the low-level process, reflecting the effectiveness of the recall strategy in guiding user to activate target memories. We define the high-level state transition distribution as \(\mathcal{P}(s_{t+1}^{h} \mid s_t^{h}, a_t^{h})\), depending on the outcomes of the low-level process, as well as updates of \(H_t\) and \(M_t\). 

\paragraph{Low Level: Cue Query Generation.} The low-level process focuses on generating memory-inspired cue query \(Q_c\) based on the selected high-level strategy pattern \(a_{t}^{h}\). Therefore, we define the low-level state as \(s_{t}^{l}=(s_{t}^{h},a_{t}^{h})\), including the high-level context and the selected strategy up at turn \(t\). 

The low-level action \(a_{t}^{l}\) represents cue query \(Q_c\), which is generated by LLM \textit{(Qwen2.5-32B-Instruct)}. The LLM itself has the characteristic of generating probabilities, where each cue query generation can be viewed as a sample drawn from a conditional probability distribution \(a_{t}^{l}\sim \mathcal{P}(a \mid s_{t}^{l})\), where \(\mathcal{P}(a \mid s_{t}^{l}) = \operatorname{LM}(Q_c \mid s_{t}^{l})\). In order to achieve the best user recall performance, we select the \(Q_c \in \mathcal{A}^l\) that maximizes the \(\mathcal{P}\). This process can be expressed as Equation~\ref{eq:3}. 
\begin{equation}
  \label{eq:3}
  a_{t}^{l} = \arg\max_{Q_c \in \mathcal{A}^l} \mathcal{P}(Q_c \mid s_t^l)
\end{equation}

\subsubsection{Recall Tree Traversal}
As shown in Figure~\ref{fig:tree}, the recall tree traversal primarily focuses on the high-level process in selecting the optimal recall strategy for each query, which contains four key steps.
\paragraph{Selection.} We define the root node as the \(Q_u\) \textit{(assuming it is none-memory-inspired)}. Each iteration starts from the root node, and then selects a child node guided by the recognized Upper Confidence Bound applied to Trees (UCT)~\cite{kocsis2006bandit} algorithm, which can balance exploration and exploitation and is expressed as Equation~\ref{eq:4}. 
\begin{equation}
  \label{eq:4}
  a = \arg\max_{a \in \mathcal{A}^h} ( \widehat{Q}_i + c \cdot \sqrt{\frac{\ln N(p)}{N(i)}} )
\end{equation}
where \(\widehat{Q}_i\) is the average reward for node \(i\), \(N(p)\) and \(N(i)\) are the visit count for parent node \(p\) and node \(i\), \(c\) is an exploration constant. The action \(a\) that maximizes the UCT value will be selected from the child node with the highest UCT value.  
\paragraph{Expansion.} When node \(i\) is not fully expanded, this phase will be executed. We use the LLM to generate low-level cue query \(Q_c\) based on the selected high-level strategy \(a_t\), expressed as \(Q_{c}=f\left ( s_t,a_t \right ) \), where \(s_t\) is the current state. After the simulated user generates response \(r_t\) based on \(Q_c\), the new state \(s_{t+1}\) is updated. The terminal state is reached when \(r_t\) exceeds the accuracy threshold or the maximum number of dialog turn is exceeded, which is expressed as Equation~\ref{eq:5}. 
\begin{equation}
  \label{eq:5}
s_{end} = 
\begin{cases} 
\text{Success, if} & \text{acc}(r_{t}, r_\text{ans}) \geq \theta_{\text{acc}}\\ 
\text{Failure,  if} & n_t \geq {N}_\text{max}
\end{cases}
\end{equation}

where \(\text{acc}(\cdot)\) is the BERTScore~\cite{zhang2019bertscore} based similarity metric, \(r_\text{ans}\) is the true answer, \(n_t\) is the number of current dialog turn. 
\paragraph{Simulation.} To estimate the future rewards, the future situations will be simulated until terminal state after expansion. To improve the effectiveness of recall guidance, we design a fine-grained instant reward mechanism by considering three dimensions: recall accuracy, recall focus, and recall depth. 
\begin{itemize}
\item \textbf{Recall Accuracy \(R_{ra}\): }Evaluation of similarity between \(r_t\) and \(r_\text{ans}\) and calculated by BERTScore. 
\item \textbf{Recall Focus \(R_{rf}\): }Evaluation of the consistency of the topic between \(r_t\) and \(r_\text{ans}\), calculated by Jaccard distance\cite{jaccard1912distribution}, expressed as Equation~\ref{eq:6}. 
\begin{equation}
  \label{eq:6}
R_{rf} = \frac{|r_t \cap r_\text{ans}|}{|r_t \cup r_\text{ans}|}
\end{equation}
\item \textbf{Recall Depth \(R_{rd}\): }Evaluation of the level of detail of the user memory reflected in the answer, which is specifically manifested in the richness of the memory elements involved in \(r_t\) and can be formulated as Equation~\ref{eq:7}. 
\begin{equation}
  \label{eq:7}
R_{rd} = \sum_{i=1}^{n} \mathbb{\textbf{1}}(e_i \in r_t)
\end{equation}
where \(e_i\) represents the \(i\)-th type of memory element in user response \(r_t\), \(n\) is the number of memory element types, including \textit{Event}, \textit{Person}, \textit{Location}, \textit{Temporal}, and \textit{Decision}, evaluated by LLM \textit{(Qwen2.5-32B-Instruct)}. The indicator function \textbf{1} takes the value of $1$ when the \(e_i\) appears in \(r_t\), and $0$ otherwise. 
\end{itemize} 

To avoid getting trapped in the local optimum, we introduce an exploration factor \(\varepsilon \) to control randomness. \(\varepsilon \) is the probability of selecting a random action, while \(1-\varepsilon \) represents the probability of selecting optimal action. We start with \(\varepsilon=1 \) and gradually decrease it by \(0.05\) during the simulation. 
\paragraph{Backpropagation.}
Once the simulation reaches a terminal state \(s_{end}\), the final reward is backpropagated from the leaf node to the root node. During this process, the reward value \(Q(s_t)\) and the visit count \(N(s_t)\) of nodes will be updated, which can be expressed as Equation~\ref{eq:8} and Equation~\ref{eq:9}.
\begin{equation}
  \label{eq:8}
Q(s_t) \leftarrow Q(s_t) + \left( \gamma^{T-t} \cdot R_{\text{end}} \right)
\end{equation}
\begin{equation}
  \label{eq:9}
N(s_t)\gets  N(s_t)+1
\end{equation}
where \(\gamma\in [0,1]\) is the discount factor, representing the degree of discount on future rewards, \(T\) is the total length of the path from the root to the leaf, and \(R_{\text{end}}\) is the final reward. 

We perform \(\textrm{T}\) iterations (\(\textrm{T}=120\)) on SGR-MCTS to construct and traverse the recall tree, ultimately identifying the top-\(k\) recall paths (\(k=5\)) with the highest total value. 

\subsection{Strategy-Driven Agent Developing}

To develop the strategy-driven memory-recall agent, we need to construct the appropriate dataset to fine tune LLMs, called MemoStrategy dataset. In particular, based MCTS-enhanced strategy exploration, we record the high-level action sequences, and low-level strategy responses corresponding to the user's original query as the corpora in MemoStrategy dataset. In the MemoStrategy dataset,each sample contains three core elements: user original query \(Q_u\), memory strategy \(s_i\), and cue query \(Q_c\). Here, \(Q_u\) represents the user query regarding forgetfulness, and \(s_i\) is determined based on the forgetfulness type identified from \(Q_u\), guiding the subsequent generation of the \(Q_c\) using the relevant \(s_i\). We collect a total of 5805 data samples, which are divided into training and testing sets, 5200 and 605 respectively. This instruction-tuning dataset includes three fields: \texttt{Instruction} is the task description and provides brief definitions for memory-recall strategies. \texttt{Input} is the user original query. \texttt{Output} represents the recommended strategy and the predicted cue query response. Details are shown in Appendix C.

We fine-tune LLMs using the MemoStrategy dataset and leverage the function calling capability of LLMs to obtain structured outputs. Based on this, we further develop the LLM-based agent capable of activating user memory through memory-inspired strategies, called MemoCue. Specifically, MemoCue can transform the user original query \textit{(none-memory-inspired)} into a cue-rich one \textit{(memory-inspired)} to activate the user's memory progressively. 

\section{Experiments}
\subsection{Experimental Settings}
\paragraph{Datasets.}We evaluate our MemoCue on three representative long-term memory datasets, all of which contain the user personal memory bank, as well as questions and answers for specific memories. 

\begin{itemize}
    \item \textbf{PerLTQA}~\cite{du2024perltqa}. This dataset includes 141 user memory banks and 8,593 memory question answer pairs. To retrieve memory streams more precisely, we transform each long query entry into multiple short sentences and obtain 35,179 memory streams in the end. 
    \item \textbf{LoCoMo}~\cite{maharana2024evaluating}. LoCoMo is a dataset which comprises 50 high-quality ultra-long conversations, where each conversation contains approximately 300 turns and 9,000 tokens. LoCoMo offers the rich personal descriptions to ensure the coherence and long-term consistency of conversations. 
    \item \textbf{MemoryBank}~\cite{zhong2024memorybank}. MemoryBank utilizes a dialogue memory storage system in the experiment created over 10 days by 15 virtual users, each possessing distinct personalities. Researchers then create 194 probing questions to assess whether the model could generate relevant responses and recall memories effectively. 

\end{itemize}

\begin{table*}[ht]
\small
\centering
\captionsetup{justification=justified, singlelinecheck=false, skip=5pt,belowskip=5pt}
\caption{Performance comparison of different open-source models and settings. The best results are highlight \textbf{in bold} and the second-best results are \underline{underlined} for three datasets, while our results are marked with shading. }
\begin{tabular}{p{3.5cm}p{2.5cm}p{2cm}p{2cm}p{2cm}p{2cm}}
\toprule[1.0pt]
\textbf{Model} & \textbf{Setting
} & \textbf{PerLTQA} & \textbf{LoCoMo} & \textbf{MemoryBank} & \textbf{Average} \\ \midrule[0.75pt]
\multirow{5}{*}{\textbf{Qwen2.5-14B-Instruct}} 
& Zero-shot & $\underline{71.32}$ & $68.04$ & $66.18$ & $\underline{68.51}$  \\  
& Few-shot & $67.94$ & $68.32$ & $66.30$ & $67.52$  \\ 
& Zero-shot CoT & $69.78$ & $\underline{70.02}$ & $64.13$ & $67.98$ \\   
& Few-shot CoT & $68.77$ & $69.64$ & $\underline{66.62}$ & $68.34$ \\   
&
\textbf{MemoCue} & $\bf{78.15}$ & $\bf{77.89}$ & $\bf{72.42}$ & $\bf{76.15}$ \\ 
\midrule
\multirow{5}{*}{\textbf{Qwen2.5-7B-Instruct}} 
& Zero-shot & $69.33$ & $65.00$ & $64.56$ & $66.30$ \\   
& Few-shot & $64.30$ & $66.26$ & $66.27$ & $65.61$ \\   
& Zero-shot CoT & $\underline{69.88}$ & $67.43$ & $67.11$ & $68.14$ \\   
& Few-shot CoT & $68.88$ & $\underline{68.85}$ & $\underline{67.61}$ & $\underline{68.45}$ \\   
& 
\textbf{MemoCue} & $\bf{76.32}$ & $\bf{73.59}$ & $\bf{70.22}$ & $\bf{73.38}$ \\ \midrule
\multirow{5}{*}{\textbf{Yi-9B}} 
& Zero-shot & $\underline{70.86}$ & $65.52$ & $64.43$ & $66.94$ \\   
& Few-shot & $68.00$ & $67.51$ & $65.13$ & $66.88$ \\   
& Zero-shot CoT & $68.59$ & $\underline{70.50}$ & $65.29$ & $68.13$ \\   
& Few-shot CoT & $67.74$ & $69.75$ & $\underline{66.94}$ & $\underline{68.14}$ \\   
& 
\textbf{MemoCue} & $\bf{73.58}$ & $\bf{75.71}$ & $\bf{70.07}$ & $\bf{73.12}$ \\ \midrule
\multirow{5}{*}{\textbf{Mistral-7B-Instruct}}
& Zero-shot & $58.45$ & $59.13$ & $62.74$ & $60.11$ \\   
& Few-shot & $60.32$ & $60.67$ & $62.88$ & $61.29$ \\  
& Zero-shot CoT & $62.87$ & $64.11$ & $63.41$ & $63.46$ \\   
& Few-shot CoT & $\underline{66.33}$ & $\underline{66.06}$ & $\underline{65.24}$ & $\underline{65.88}$ \\   
& 
\textbf{MemoCue} & $\bf{69.21}$ & $\bf{72.01}$ & $\bf{70.93}$ & $\bf{70.72}$ \\ \midrule
\multirow{5}{*}{\textbf{Llama-3-8B-Instruct}} 
& Zero-shot & $63.78$ & $\underline{67.60}$ & $65.31$ & $65.56$ \\   
& Few-shot & $60.46$ & $63.15$ & $64.82$ & $62.64$ \\   
& Zero-shot CoT & $\underline{68.43}$ & $66.81$ & $\underline{69.30}$ & $\underline{68.18}$ \\   
& Few-shot CoT & $67.33$ & $66.45$ & $70.31$ & $68.03$ \\   
& 
\textbf{MemoCue} & $\bf{72.65}$ & $\bf{73.33}$ & $\bf{71.24}$ & $\bf{72.41}$ \\ \bottomrule
\end{tabular}
\label{tab:performance}
\end{table*}

\paragraph{Implementation Details.}We perform the fine tuning on five open-source LLMs with our constructed MemoStrategy dataset, which includes Qwen2.5-14B-Instruct, Qwen2.5-7B-Instruct~\cite{yang2024qwen2}, LlaMA-3-8B-Instruct~\cite{meta2024introducing}, Mistral-7B-Instruct~\cite{jiang2023mistral} and Yi-9B~\cite{young2024yi}. During the phase, we use a batch size of 8 per GPU and a learning rate of 2e-5 for all open-source LLMs. We train our model on 8 Nvidia Tesla A100-80GB GPUs with LoRA~\cite{hulora} on 5 epochs.

\paragraph{Baselines.} We compare MemoCue with competitive open-source models and closed-source models enhanced by two types of In-Context Learning (ICL) methods: 1) \textit{Standard Prompting,} including Zero-shot Prompting~\cite{wei2021finetuned} and Few-shot Prompting, which provides the model with a direct prompt to perform a task. In Zero-shot Prompting, the model is directly generated with relying solely on its pre-trained knowledge. In Few-shot Prompting, a small number of demonstrations are provided to instruct the model to obtain the better performance. 2) \textit{Chain-of-Thought Prompting}~\cite{wei2022chain}, including Zero-shot CoT~\cite{kojima2022large}, and Few-shot CoT~\cite{wei2022chain}, which encourages the model to break down complex tasks into intermediate reasoning steps, thereby improving the transparency of the model’s reasoning process and the accuracy of complex reasoning tasks. 
\begin{itemize}
    \item \textbf{Open-Source Models}. In order to enhance the reproducibility of exprimental results, we choose five open-sourced billion-level LLMs, including Qwen2.5-14B-Instruct, Qwen2.5-7B-Instruct, LlaMA-3-8B-Instruct, Mistral-7B-Instruct and Yi-9B.
    \item \textbf{Closed-Source Models}. In order to comprehensively compare the experimental results, we choose gpt-3.5-turbo, gpt-4~\cite{achiam2023gpt}, o1-preview~\cite{zhong2024evaluation} and gpt-4o~\cite{hurst2024gpt} as closed-source models in baselines. 
\end{itemize}

Note that, due to the lack of real-world human memory recall datasets, it is difficult to directly evaluate the end-to-end recall success performance in this work. Instead, we demonstrate the performance of proposed system at the following three aspects: (1) How much can MemoCue improve recall effectiveness, excluding cases where responses merely repeat the original query? (2) How much can MemoCue improve the performance of response in terms of fine-grained content? (3) How much can MemoCue improve the performance of response in real-user experiences?

\subsubsection{Evaluation Metrics.}
We consider that the agent can guide users to recall the memory by generating memory-inspired cue queries. The quality of generated memory-inspired cue queries determines the memory recalling performance. 
On one hand, the generated cue query should avoid being too similar to the original query to prevent invalid repetition. On the other hand, it should not stray too far from the topic to avoid affecting the accuracy of the answer. Therefore, we present the metrics for the following three evaluation methods, to evaluate the quality of generated cue queries. 

\paragraph{Automatic Evaluation.} In order to evaluate the effectiveness of MemoCue, we design an evaluation metric \textbf{B}alance of \textbf{R}ecall \textbf{S}core \textbf{(BRS)} to balance ``\textit{the difference between the original query and the cue query}'' and ``\textit{the accuracy of memory retrieval}''. Specifically, the former is used as a penalty factor and the latter is used as a reward factor, which is mathematically expressed as Equation~\ref{eq:10}.
\begin{equation}
  \label{eq:10}
\text{BRS} = \frac{\text{Acc}(r_{res,i}, r_{ans,i})}{1 + \alpha \cdot \text{Sim}(Q_{u,i}, Q_{c,i})} 
\end{equation}
where \(\text{Acc}(\cdot, \cdot)\) is the accuracy of response calculated by BERTScore, \(\text{Sim}(\cdot, \cdot)\) is the cosine similarity of two queries. Here, \(\alpha\) is a hyperparameter used to adjust the penalty intensity, which is set to $0.3$ here.
    
\paragraph{LLM Evaluation.} To achieve a more comprehensive assessment of recall performance, we also employ LLM-based evaluation. This provides the assessment across multiple dimensions, including \textit{recall accuracy, recall coherence, recall logicality, recall inspiration,} and \textit{human-likeness} of the generated cue queries. In particular, we employ OpenAI o1-Preview, the most powerful inference model yet, to evaluate the quality of cue queries from the above dimensions using few-shot prompts. In order to alleviate the position bias of LLM evaluation, we employ the balanced position calibration (BPC) strategy~\cite{wang2023large}. To ensure the reproducibility of the results, we set the temperature as $0$. 

\paragraph{Human Evaluation.} To evaluate the performance of MemoCue in real-world scenarios, we employ human evaluators to assess the responses generated by MemoCue. To ensure the practicality of MemoCue across different age groups, we employ 9 evaluators from three age ranges: under 30, 30-50, and over 50. Specifically, we first provide the evaluators with specialized training. Then, we present 30 pairs of original queries alongside responses generated by both LLM and MemoCue, with the order randomized. Evaluators rank the response pair with the following dimensions: \textit{recall inspiration}, \textit{response intelligence}, and \textit{cue consistency}. 
\subsection{Experimental Results}
\begin{figure}[tb]
\small
    \centering
    \captionsetup{justification=justified, singlelinecheck=false, skip=6pt,belowskip=5pt}
    \includegraphics[width=0.38\textwidth]{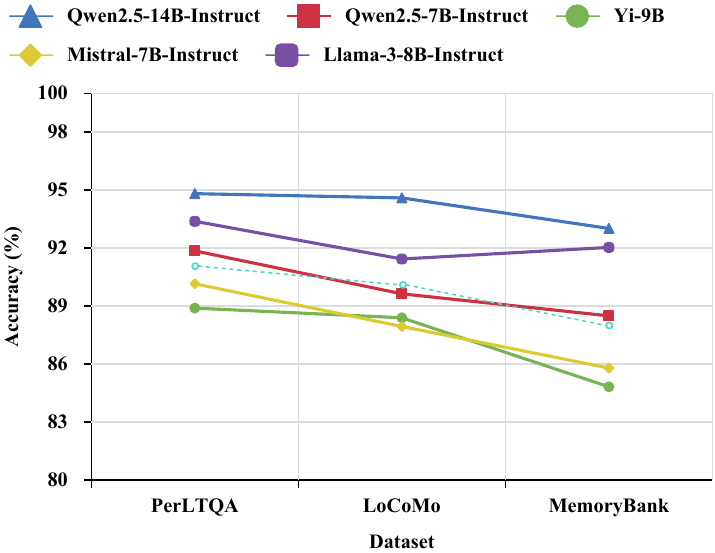}
    \caption{The evaluation results of the accuracy of MemoCue on memory-recall strategy prediction for different LLMs. The dashed line represents the average accuracy within each dataset. }
    \label{fig:eval-acc}
\end{figure}

We first conduct a series of experiments to investigate the accuracy of MemoCue on memory-recall strategy prediction, which are fine-tuned on different LLMs across three long-term memory related datasets. It can be seen from Figure~\ref{fig:eval-acc} that MemoCue can achieve the average accuracy of \textbf{90.36\%} over considered five LLMs. Based on this, we conduct further evaluations. 
\paragraph{Automatic Evaluation Results.} Table~\ref{tab:performance} shows the main evaluation results of our experiments. It can be seen that MemoCue achieves the highest BRS among all considered mainstream open-source LLMs across three representative datasets. For example, MemoCue fine-tuned based on Qwen2.5-14B-Instruct improves the BRS by $\textbf{11.24\%}$ (compared to zero-shot CoT) on LoCoMo dataset, $\textbf{9.58\%}$ (compared to zero-shot) on PerLTQA dataset, and $\textbf{8.71\%}$ (compared to few-shot CoT) on MemoryBank dataset, demonstrating an enhanced ability to generate cue queries that balance the novelty and response accuracy.

To more convincingly demonstrate the superiority of our MemoCue, we compare MemoCue (fine-tuned on Qwen2.5-14B-Instruct) with the leading closed-source LLMs enhanced with zero-shot CoT and few-shot CoT prompting. As shown in Table~\ref{tab:number}, MemoCue outperforms these models in terms of BRS.

\begin{table}[h]
\small
\centering
\captionsetup{justification=justified, singlelinecheck=false, skip=5pt,belowskip=5pt}
\caption{Performance comparison of different closed-source models.}
\begin{tabular}{cccc}
\toprule[1.0pt]
\textbf{Model} & \textbf{Setting} & \textbf{PerLTQA} & \textbf{LoCoMo} \\ \midrule[0.75pt]
\multirow{2}{*}{\textbf{gpt-3.5-turbo}}  
& Zero-shot CoT & $65.43$ & $67.82$ \\   
& Few-shot CoT & $66.34$ & $68.11$ \\     
 \midrule
\multirow{2}{*}{\textbf{gpt-4}}   
& Zero-shot CoT & $68.54$ & $68.02$ \\   
& Few-shot CoT & $69.76$ & $69.90$ \\     
 \midrule
\multirow{2}{*}{\textbf{gpt-4o}} 
& Zero-shot CoT & $69.32$ & $68.81$ \\   
& Few-shot CoT & $70.11$ & $69.53$ \\    
 \midrule
\multirow{2}{*}{\textbf{o1-preview}} 
& Zero-shot CoT & $70.81$ & $68.14$ \\   
& Few-shot CoT & $71.32$ & $69.94$ \\   
\midrule
& 
\textbf{MemoCue} & $\bf{76.32}$ & $\bf{73.59}$ \\  
 \bottomrule
\end{tabular}
\label{tab:number}
\end{table}

\begin{figure}[htbp]
\small
    \centering
    \captionsetup{justification=justified, singlelinecheck=false, skip=6pt,belowskip=5pt}
    \includegraphics[width=0.38\textwidth]{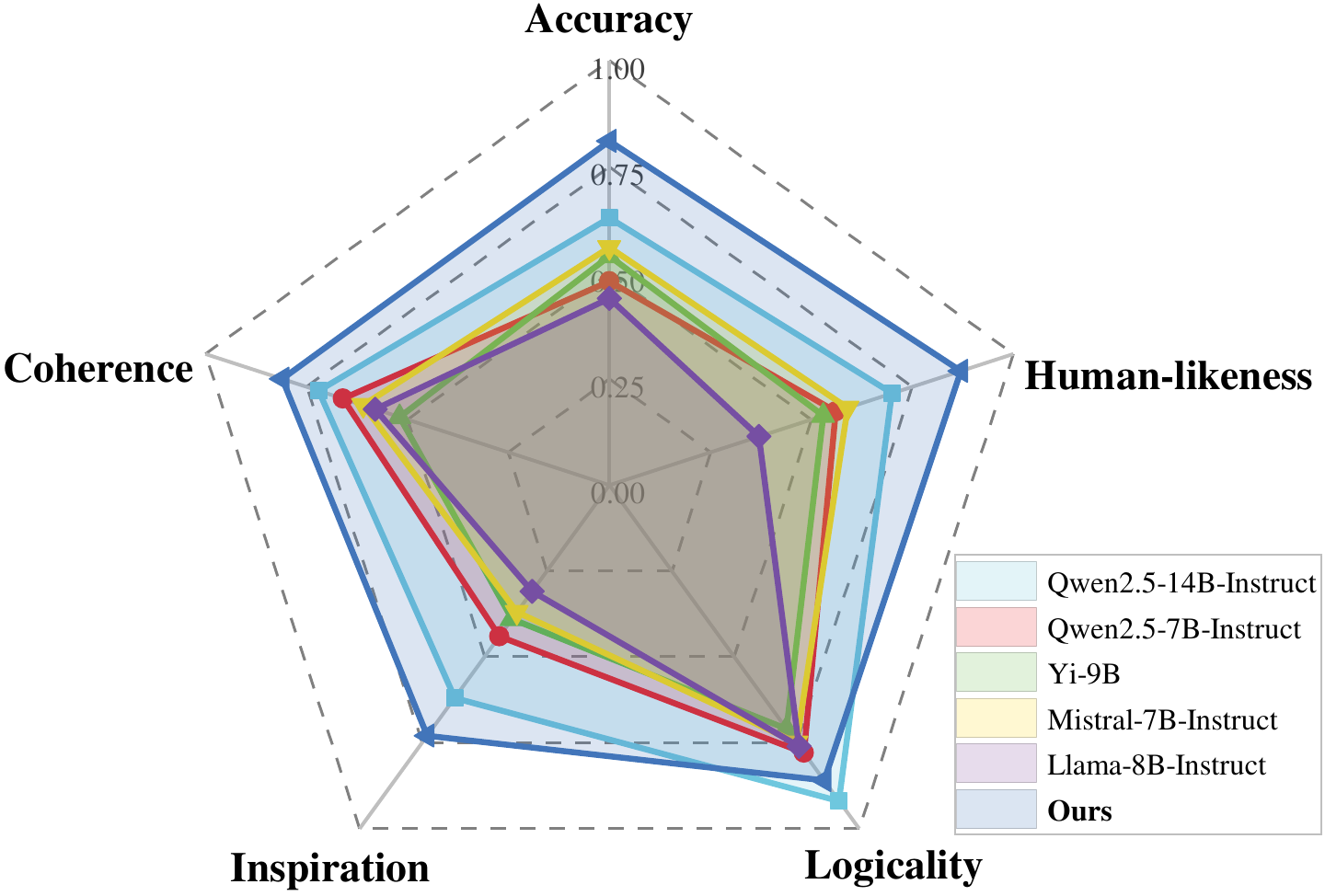}
    \caption{The LLM-based evaluation results. }
    \label{fig:eval-o1}
\end{figure}

\paragraph{LLM Evaluation Results.} Figure~\ref{fig:eval-o1} shows the LLM-based performance evaluation. It can be seen that MemoCue outperforms baselines generally. For example, MemoCue improves the recall inspiration of response expression by $\textbf{17.74\%}$.
For the metric of recall logicality, the performance of MemoCue is slightly worse than that of Qwen2.5-14B-Instruct model. This is because the larger-scale parameters enable the model to capture richer logical relationships. 

\paragraph{Human Evaluation Results.} Table~\ref{tab:open} shows the final win rate distribution on human evaluation. The results show that MemoCue generally outperforms competitive closed-source models in terms of recall inspiration, response intelligence, and cue consistency. 

\begin{table}[htbp]
\small
\captionsetup{justification=justified, singlelinecheck=false, skip=5pt,belowskip=5pt}
\caption{Win Rate of MemoCue over competitive closed-source models with HumanEval serving as evaluator.}
\label{tab:open}
\centering
\begin{tabular}{lcccc}
\toprule
\textbf{Evaluator} & v.s. \textbf{gpt-4} & v.s. \textbf{gpt-4o} & v.s. \textbf{o1-preview}\\
\midrule
HumanEval & 83\% & 85\% & 78\% \\
\bottomrule
\end{tabular}
\end{table}

\subsection{Ablation Study}
We also conduct a series of ablation studies to investigate the impact of each module in our proposed framework, as shown in Table~\ref{tab:ablation}. The whole framework, incorporating both 5W recall map and SGR-MCTS, achieves the best performance of all three datasets. Removing 5W recall map generally leads to a score drop by $3$$\sim$$4$, indicating that distinguishing different scenarios and using corresponding strategies can facilitate the generation of more effective responses. Removing SGR-MCTS module decreases the score by around $4$$\sim$$5$, showing that SGR-MCTS can generate high-quality responses by exploiting the simulated user feedback and reward mechanism. 

\begin{table}[htbp]
\small
    \centering
    \captionsetup{justification=justified, singlelinecheck=false, skip=5pt,belowskip=5pt}
        \caption{The evaluation results of ablation study.}
    \resizebox{1.0\linewidth}{!}{
    \begin{tabular}{lrrrr}
    \toprule 
           \textbf{Method}& \bf{PerLTQA}&\textbf{LoCoMo} & \textbf{MemoryBank} & \textbf{Average} \\
         \midrule
         \textbf{MemoCue} &$\bf{78.15}$ &$\bf{77.89}$&$\bf{72.42}$&$\bf{76.15}$ \\
         \ \textit{w/o} 5W &{$\downarrow3.54$} &{$\downarrow3.27$}&{$\downarrow3.02$}&{$\downarrow3.28$} \\
         \ \textit{w/o} MCTS&{$\downarrow4.98$} &{$\downarrow5.13$}&{$\downarrow4.78$}&{$\downarrow4.96$} \\
         \ \textit{w/o} 5W\&MCTS &{$\downarrow6.83$} &{ $\downarrow9.85$}&{$\downarrow6.24$}&{$\downarrow7.64$} \\
        \bottomrule
    \end{tabular}}
    \label{tab:ablation}
\end{table}

\subsection{Effect of Number of Iterations in SGR-MCTS}
\begin{figure}[htbp]
    \centering
    \captionsetup{justification=justified, singlelinecheck=false, skip=6pt,belowskip=3pt}
    \includegraphics[width=0.34\textwidth]{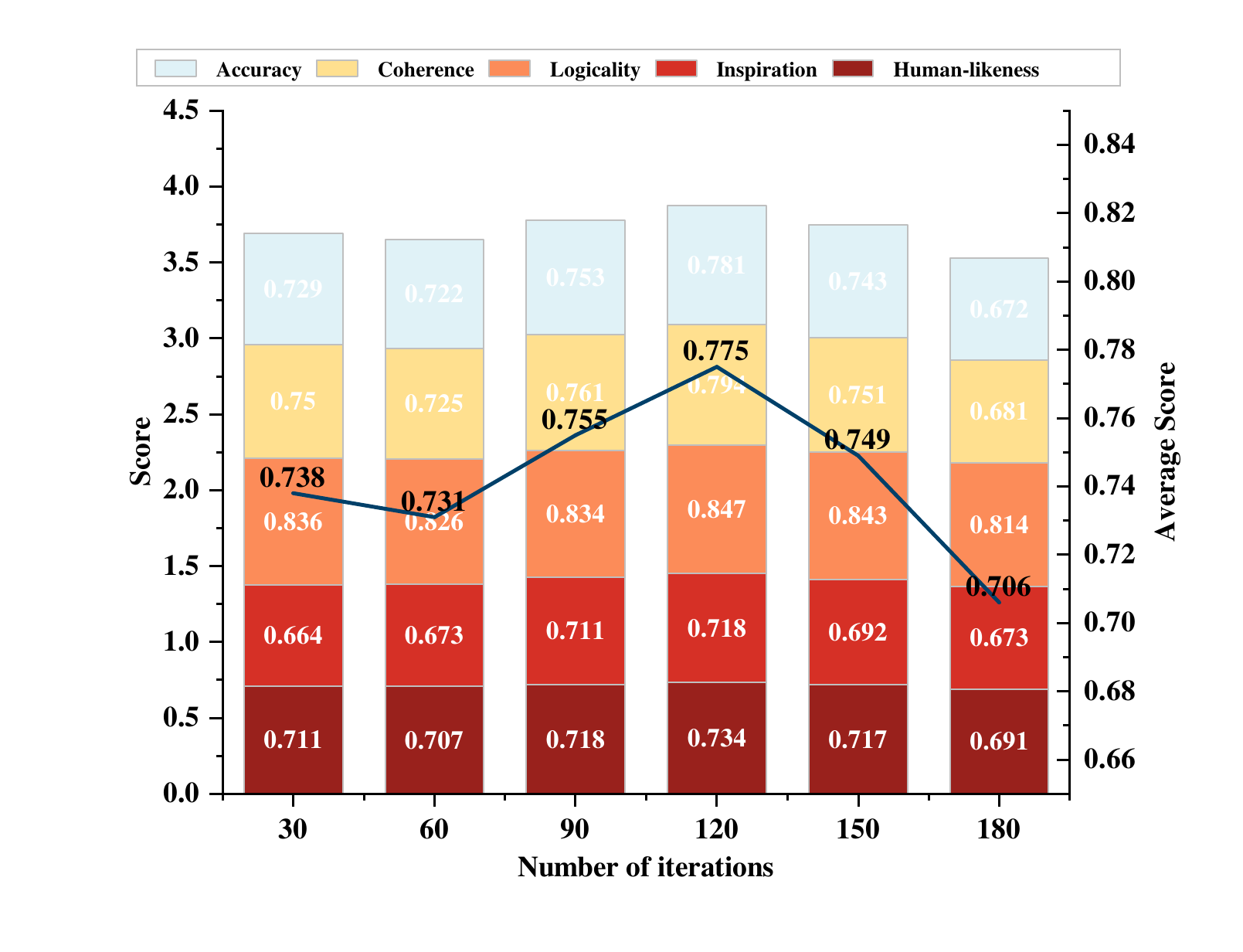}
    \caption{Impact of specifying the number of iterations in SGR-MCTS algorithm on PerLTQA dataset. }
    \label{fig:iteration}
\end{figure}

We further explore the effect of different number of iterations $\textrm{T}$ in Figure~\ref{fig:iteration}, a crucial hyperparameter influencing both the accuracy of the search results and the computational resource consumption. As $\textrm{T}$ increases, the computational time cost grows. Therefore, we set $\textrm{T}$ as $30$, $60$, $90$, $120$, $150$, and $180$ to explore the best. With an increasing number of $\textrm{T}$, the average score gradually rises and then drops. The peak score occurs when $\textrm{T}=120$. 

\section{Conclusion}
In this paper, we formulate a novel SGR problem, which aims to empower LLM-based agent for human memory recall. Specifically, we propose a Recall Router framework based on 5W recall map and SGR-MCTS algorithm. We also design a reward mechanism to explore optimal recall strategies. In addition, we construct the MemoStrategy dataset and obtain MemoCue for memory-recall guidance. Furthermore, we define an evaluation metric and conduct extensive experiments on three representative datasets compared with competitive LLM-based methods. The evaluation results show that our MemoCue has improved performance in assisting human memory recall. In the future, we will explore adaptive iteration strategies to improve scalability and conduct a pilot study with real users to further validate the practical utility of our system.  








\bibliography{main}
\renewcommand{\thesection}{\Alph{section}}
\newpage
\setcounter{section}{0}
\begin{figure*}[h]
\centering
  \includegraphics[width=0.9\linewidth]{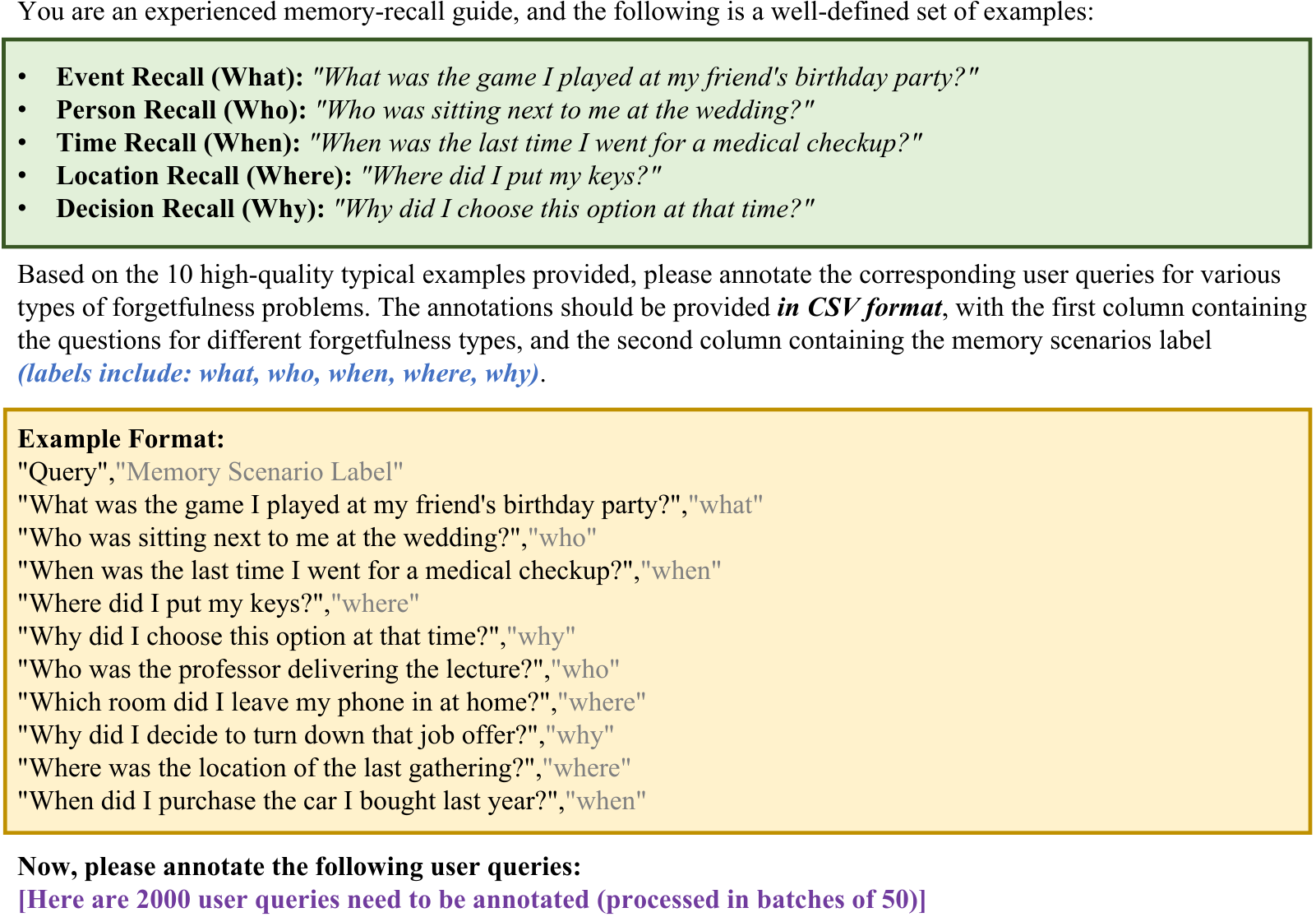}
  \caption{The prompt template of user query map to forgetting scenario.}
  \label{fig:prompt}
\end{figure*}
\section{Definitions of Memory-Recall Strategies}
We define 15 memory-recall strategies under five typical forgetting scenarios based on human memory theory. Details of the description of the 15 memory-recall strategies are shown in Table~\ref{tab:strategies}. 

\section{Details of User Query Classification by fine-tuning ReBERTa Model}
We employ GPT-3.5-turbo and utilized few-shot prompting to annotate $2,000$ data samples, with the prompt template shown in Figure~\ref{fig:prompt}. These samples were manually verified and then used as training data to fine-tune a pre-trained RoBERTa model. The fine-tuning is performed on an Nvidia Tesla A100-80GB GPU, with $500$ epochs, a batch size of $32$, and the key parameters detailed in Table~\ref{tab:parameter}.

\begin{table}[h]
\centering
\caption{key parameters set in RoBERTa model}
\begin{tabular}{p{5cm}p{2cm}}
\toprule
\textbf{Parameter Name} & \textbf{Value} \\ \midrule
model\_type & roberta \\
num\_hidden\_layers & 24 \\
hidden\_size & 1024 \\
num\_attention\_heads & 16 \\
intermediate\_size & 4096 \\
hidden\_act & gelu \\
attention\_probs\_dropout\_prob & 0.1 \\
hidden\_dropout\_prob & 0.1 \\
initializer\_range & 0.02 \\
layer\_norm\_eps & 1e-5 \\
vocab\_size & 50265 \\
max\_position\_embeddings & 514 \\
type\_vocab\_size & 1 \\
pad\_token\_id & 1 \\
bos\_token\_id & 0 \\
eos\_token\_id & 2 \\ \bottomrule
\end{tabular}
\label{tab:parameter}
\end{table}

\begin{table*}[h]
\centering
\caption{The description of 15 strategies on five forgetting scenarios. }
\begin{tabular}{p{2.5cm}|p{4cm}|p{10cm}}
\toprule
\textbf{Scenario} & \textbf{Strategies} & \textbf{Description} \\ \midrule
\multirow{9}{*}{Event (What)} 
    & \multirow{3}{*}{Scenario Reconstruction} & \textit{Activate and enhance memory by simulating or recreating relevant situations, helping user reconstruct memory scenes by recalling details such as surroundings, emotions, time, space, and other contextual elements.} \\    \cmidrule{2-3}
    & \multirow{3}{*}{Interpersonal Interaction} & \textit{Help user recall specific details of their interactions with others by triggering memory cues linked to particular people or social situations, using both verbal and visual information.} \\    \cmidrule{2-3}
    & \multirow{3}{*}{Sensory Activation} & \textit{Activate memory by using sensory cues like sight, sound, and smell, particularly when these stimuli are closely linked to the memory of a specific event or situation.} \\
    \midrule
\multirow{9}{*}{Person (Who)} 
    & \multirow{3}{*}{Appearance Clues} & \textit{Activate memory by using external features of another person, such as facial expressions and body language, to help recall memories related to that person through recognizing and interpreting their behavior.} \\\cmidrule{2-3}
    & \multirow{3}{*}{Role Connection} & \textit{Recall memories linked to an individual based on the user's specific relationship with them in social interactions, such as teacher-student, superior-subordinate, family member, friend, coworker, etc.} \\\cmidrule{2-3}
    & \multirow{3}{*}{Emotion Trigger} & \textit{Use emotional responses tied to specific memories, such as happiness, sadness, anger, or surprise, to activate memory by leveraging the strong connection between emotions and memories.} \\
    \midrule
\multirow{9}{*}{Location (Where)} 
    & \multirow{3}{*}{Multiple Associations} & \textit{Establish connections between multiple memory points to guide the user, using objects, specific locations, or unrelated events to broaden associative thinking and help recall.} \\\cmidrule{2-3}
    & \multirow{3}{*}{Immersive Recall} & \textit{Guide the user to re-experience a situation by using landmarks, paths, or scene details to help recall specific events or experiences from that time and place.} \\\cmidrule{2-3}
    & \multirow{4}{*}{Spatial Cues} & \textit{Use spatial information in the environment to trigger memory, helping the user more accurately recall events and details by recalling the layout of objects, buildings, directions, distances, or specific landmarks at a location.} \\
    \midrule
\multirow{8}{*}{Temporal (When)} 
    & \multirow{2}{*}{Timeline Rewind} & \textit{Activate memory through simulated time travel, allowing user to retrace past events by recreating or simulating a sequence of occurrences.} \\\cmidrule{2-3}
    & \multirow{2}{*}{Key Milestones} & \textit{Guide user to identify key events or turning points in memory, using them as anchors to help locate and recall information along the timeline.} \\\cmidrule{2-3}
    & \multirow{4}{*}{Routine Pattern} & \textit{Use regular and habitual behaviors in daily life to trigger memories, linking the user’s repeated actions at specific times, places, or situations to memory cues, helping them recall events and details associated with those behaviors.} \\\midrule
\multirow{9}{*}{Decision (Why)} 
    & \multirow{4}{*}{Background motivation} & \textit{Explore the internal psychological drivers behind the user's actions in a given situation, including intrinsic needs like self-concept, values, and goals, as well as extrinsic factors such as rewards, punishments, or societal expectations.} \\\cmidrule{2-3}
    & \multirow{3}{*}{Option Comparison} & \textit{Involve comparing different options during decision-making, helping the user recall their decision-making process by evaluating the pros and cons of each option.} \\\cmidrule{2-3}
    & \multirow{2}{*}{Experience Support} & \textit{Assist in remembering the reasons for current decisions by guiding the user to analyze past experiences.} \\ \bottomrule
\end{tabular}
  \label{tab:strategies}
\end{table*}

\begin{figure*}[h]
\centering
  \includegraphics[width=\linewidth]{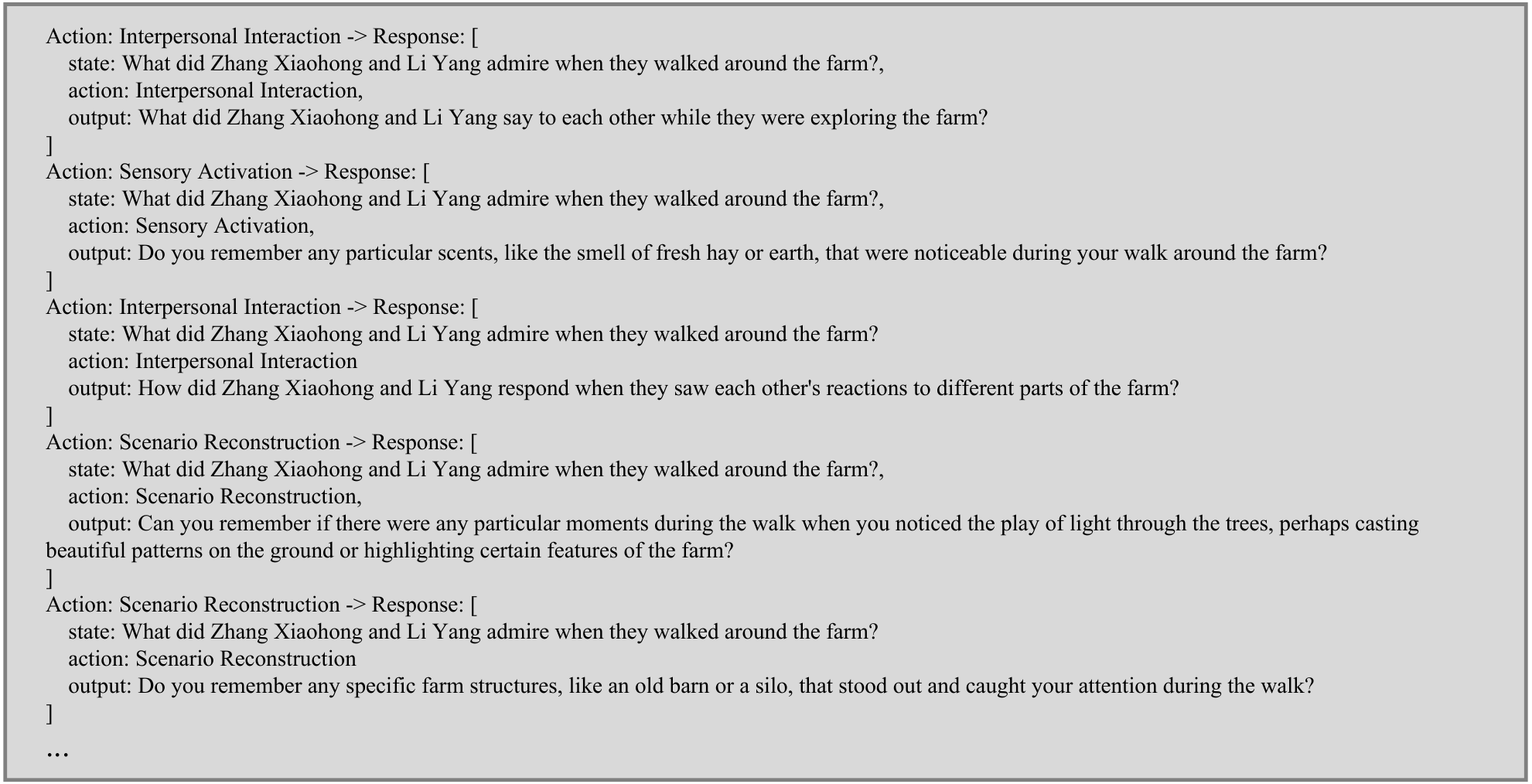}
  \caption{The original data generated by the SGR-MCTS algorithm.}
  \label{fig:origin}
\end{figure*}

\begin{figure*}[h]
\centering
  \includegraphics[width=\linewidth]{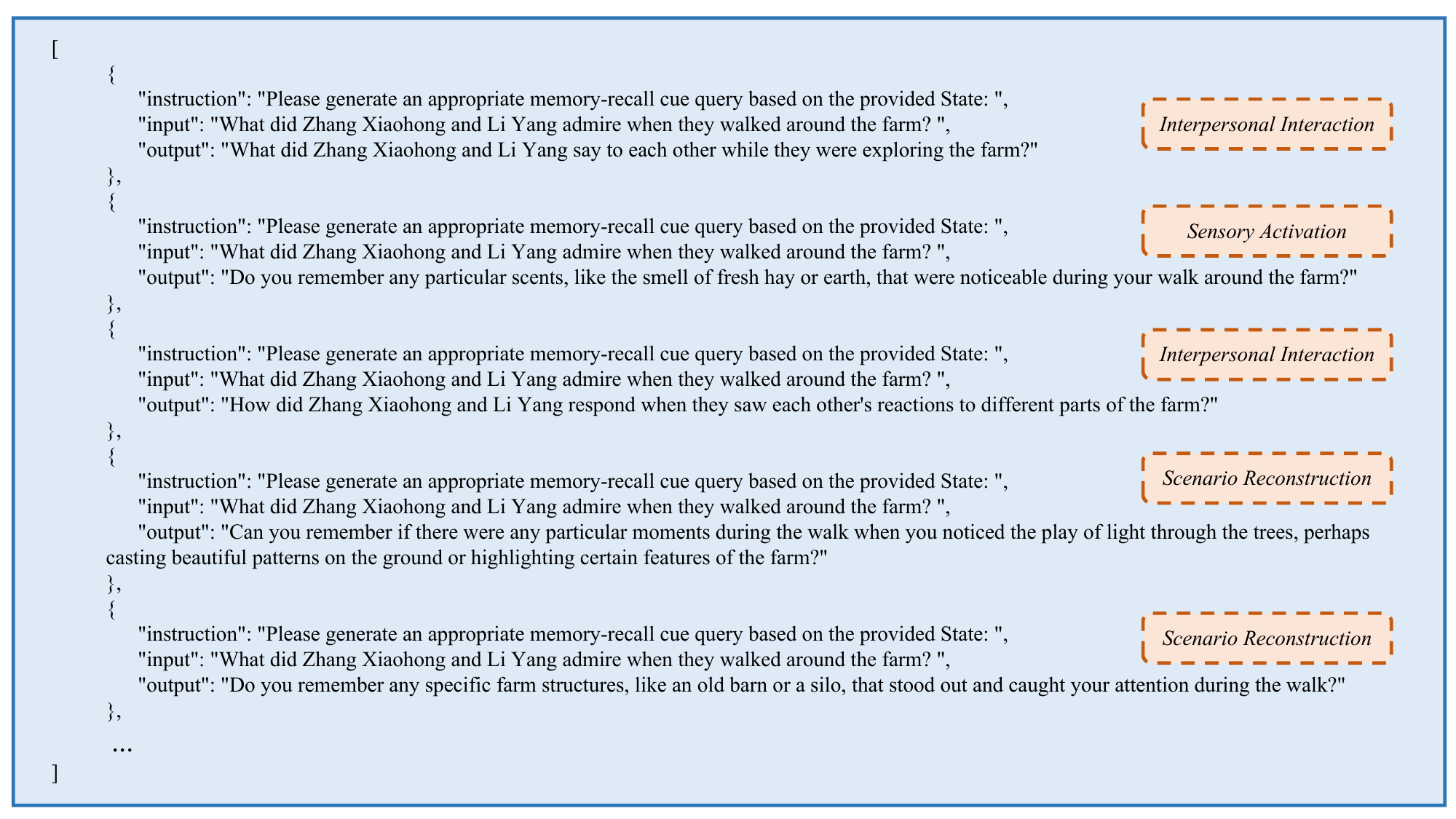}
  \caption{Examples of the Instruction-tuning MemoStrategy dataset.}
  \label{fig:dataset}
\end{figure*}

\section{Details of MemoStrategy Dataset}
\subsection{Construction of MemoStrategy Dataset}
\paragraph{Strategic Corpus Collection} We employ SGR-MCTS algorithm to select appropriate memory-recall strategies and generate corresponding cue queries, as shown in Figure~\ref{fig:origin}. First, by analyzing the user original query, we identify the type of forgetfulness and then choose a suitable memory-recall strategy to generate cue query aimed at activating user's memory. For instance, when the original query is \textit{``When did I take my medication?''}, which is related to person, a cue query such as \textit{``What time do you usually take your medication?''} by using the \textit{``routine pattern''} memory-recall strategy will be generated accordingly, which activate memory from the perspective of daily habits. The user response will serve as feedback, and its effectiveness is quantified through an immediate reward mechanism. In this way, we can efficiently simulate the memory-recall guidance effect and provide strong support for the optimization of memory-recall strategies data.
\paragraph{Data Quality Optimization.}To ensure high standards of data quality, we conduct strict screening and quality control on all memory-recall corpora. By comparing user responses and true answers, we delete invalid samples that failed to successfully activate user memory-recall, that is, scenarios where user responses were unclear or failed to provide valid information. For example, users fail to answer key information in the response (such as dates, people, etc.) in some cases, which are marked as failed samples and remove from the dataset.
\paragraph{Formatting and Instruction Creation.}In constructing the MemoStrategy dataset, we first standardize and format the data. Each sample contains three core elements: user original query \(Q_u\), memory strategy \(s_i\), and cue query \(Q_c\). \(Q_u\) represents the user query regarding forgetfulness, while \(s_i\) is determined based on the forgetfulness type identified from \(Q_u\), guiding the subsequent generation of the \(Q_c\) using the relevant \(s_i\). This instruction-tuning dataset includes three fields: \texttt{Instruction} is the task description and provides definitions for various memory-recall strategy types. \texttt{Input} is the user original query and the recommended strategy. \texttt{Output} represents the predicted cue query response. We collect a total of 4805 data samples, which are divided into training and testing sets, 4500 and 305 respectively.
\subsection{Examples of MemoStrategy Dataset}
Details of the Instruction-tuning MemoStrategy dataset are shown in Figure~\ref{fig:dataset}, which takes the user query \textit{``What did Zhang Xiaohong and Li Yang admire when they walked around the farm?''} for example.

\end{document}